\documentclass[copyright,noncommercial]{eptcs}
\pdfoutput=1
\providecommand{\volume}{5}
\providecommand{\anno}{2009}
\providecommand{\firstpage}{81}
\providecommand{\eid}{7}
\providecommand{\event}{Y. Deville and C. Solnon (Eds): Sixth International Workshop\\
 on Local Search Techniques in Constraint Satisfaction (LSCS'09).}
\usepackage{breakurl}        

\usepackage{graphicx}
\usepackage{moreverb}
\usepackage{url}
\usepackage[lined,algonl,boxed]{algorithm2e}
\usepackage{color}

\usepackage{float}
\newfloat{tab}{thp}{lop}
\floatname{tab}{Table}

\floatstyle{ruled}
\newfloat{stmt}{thp}{lop}
\floatname{stmt}{Statement}

\newcommand\comet{\textsc{Comet }}

\title{Sonet Network Design Problems}

\author{Marie Pelleau$^{\dag,\ddag}$ \qquad\qquad Pascal Van Hentenryck$^\ddag$ \qquad\qquad Charlotte Truchet$^\dag$
\email{marie.pelleau@univ-nantes.fr \qquad\qquad pvh@cs.brown.edu \qquad\qquad Charlotte.Truchet@univ-nantes.fr}\\
\institute{
\begin{tabular}{c c c}
$\dag$ Universit\'e de Nantes & \qquad\qquad\qquad & $\ddag$ Brown University\\
D\'epartement informatique & &  Computer Science\\
2 rue de la Houssini\`ere & &  115 Waterman Street\\
44322 Nantes cedex 3 & &  Providence, RI 02912\\
France & &  USA\\
\end{tabular}}
}

\def\titlerunning{Sonet Network Design Problems}
\def\authorrunning{M. Pelleau \& P. Van Hentenryck \& C. Truchet}
\begin{document}
\maketitle

\begin{abstract}
This paper presents a new method and a constraint-based objective function to solve two problems related to the design of optical telecommunication networks, namely the Synchronous Optical Network Ring Assignment Problem (SRAP) and the Intra-ring Synchronous Optical Network Design Problem (IDP). These network topology problems can be represented as a graph partitioning with capacity constraints as shown in previous works. We present here a new objective function and a new local search algorithm to solve these problems. Experiments conducted in \comet allow us to compare our method to previous ones and show that we obtain better results.
\end{abstract}

\section{Introduction \label{sec:intro}}

This paper presents a new algorithm and an objective function to solve two real-world combinatorial optimization problems from the field of network design. These two problems, the Synchronous Optical Network Ring Assignment Problem (SRAP) and the Intra-ring Synchronous Optical Network Design Problem (IDP), have been shown $\mathcal{NP}$-hard and have already been solved by combinatorial optimization techniques. This work extends the seminal ideas introduced by R. Aringhieri and M. Dell'Amico in 2005 in ~\cite{AD2005}.

\paragraph{}
This paper is organized as follows. In the sequel of this section we introduce the two problems we have worked on, and the local search techniques which have been used to solve them. We will also introduce the models in a constrained optimization format for the two problems.
We then present the previous works on SRAP and IDP in section \ref{sec:rwork}. Section \ref{sec:work} describes the key ingredients necessary to implement the local search algorithms. Finally, the results are shown in Section \ref{sec:results}.
 
\subsection{Optical networks topologies} 
During the last few years the number of internet based application users has exponentially increased, and so has the demand for bandwidth. To enable fast transmission of large quantities of data, the fiber optic technology in telecommunication is the current solution.

The Synchronous Optical NETwork (SONET) in North America and Synchronous Digital Hierarchy (SDH) in Europe and Japan are the standard designs for fiber optics networks. They have a ring-based topology, in other words, they are a collection of rings. 

\paragraph{Rings} Each customer is connected to one or more rings, and can send, receive and relay messages using an add-drop-multiplexer (ADM). There are two bidirectional links connecting each customer to his neighboring customers on the ring. In a bidirectional ring the traffic between two nodes can be sent clockwise or counterclockwise. This topology allows an enhanced survivability of the network, specifically if a failure occurs on a link, the traffic originally transmitted on this link will be sent on the surviving part of the ring. The volume traffic on any ring is limited by the link capacity, called $B$. The cost of this kind of network is defined by the cost of the different components used in it.

There are different ways to represent a network. In this paper, we consider two network topologies described by R. Aringhieri and M. Dell'Amico in 2005 in ~\cite{AD2005}. In both topologies the goal is to minimize the cost of the network while guaranteeing that the customers' demands, in term of bandwidth, are satisfied.

The model associated to these topologies are based on graphs.
Given an undirected graph $G = (V, E)$, $V = \{1, \dots, n\}$, the set of nodes represent the customers and $E$, the set of edges, stand for  the customers' traffic demands. A communication between two customers $u$ and $v$ corresponds to the weighted edge $(u, v)$ in the graph, where the weight $d_{uv}$ is the fixed traffic demand. Note that $d_{uv} = d_{vu}$, and that $d_{uu} = 0$.

\subsubsection{First topology (SRAP)} 
\begin{figure}[h]
 \begin{center}
  \includegraphics[scale=0.6]{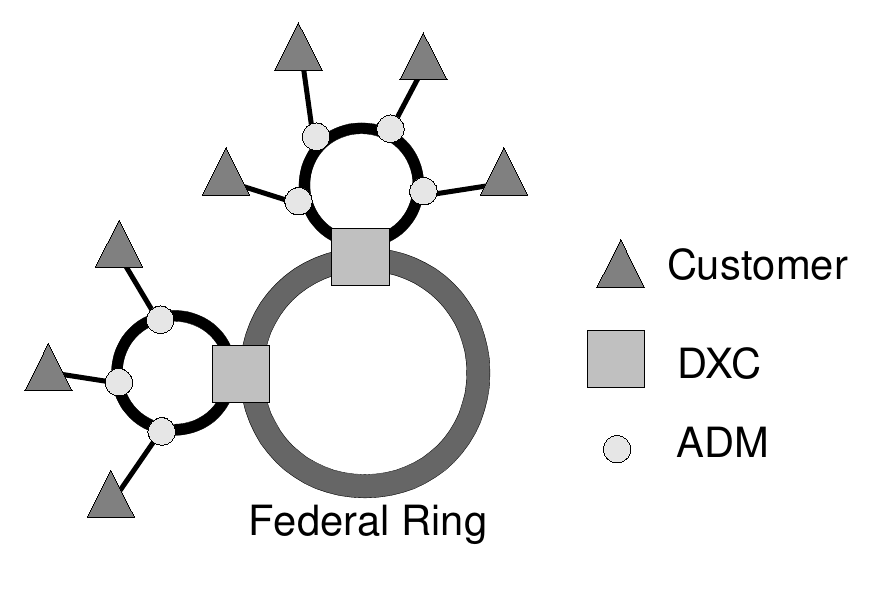}
  \caption{A SONET network with DXC. \label{fig:top1}}
 \end{center}
\end{figure}

In the first topology, each customer is connected to exactly one ring. All of these \textit{local rings} are connected with a device called digital cross connector (DXC) to a special ring, called the {\it federal ring}. The traffic between two rings is transmitted over this special ring. Like the other rings, the federal ring is limited by the capacity $B$. Because DXCs are so much more expensive than ADMs we want to have the smallest possible number of them. As there is a one-to-one relationship between the ring and the DXC, minimizing the number of rings is equivalent to minimizing the number of DXCs. The problem associated to this topology is called \textit{SONET Ring Assignment Problem} (SRAP) \textit{with capacity constraint}. Figure \ref{fig:top1} shows an example of this topology.

\paragraph{Model} This topology is modeled by a decomposition of the set of nodes $V$ into a partition, each subset of the partition representing a particular ring. Assigning a node to a subset of the partition in the model is then equivalent to assigning a customer to a ring.

Formally, let $V_1, V_2, \dots, V_k$ be a partitioning of $V$ in $k$ subsets. Each customer in the subset $V_i$ is assigned to the $i$-th local ring. As each customer is connected  with an ADM to one and only one ring, and each local ring is connected to the federal ring with a DXC, there are exactly $|V|$ AMD and $k$ DXC used in the corresponding SRAP network.

Hence, minimizing the number of rings is equivalent to minimizing $k$ subject to the following constraints:
\begin{equation}
  \sum_{u \in V_i}\sum_{v \in V, v \neq u}d_{uv} \leq B, \qquad \forall i = 1, \dots, k \label{eq:consSRAP1}
\end{equation}
\begin{equation}
  \sum_{i = 1}^{k - 1}\sum_{j = i+1}^k\sum_{u \in V_i}\sum_{v \in V_j}d_{uv} \leq B \label{eq:consSRAP2}
\end{equation}
Constraint (\ref{eq:consSRAP1}) imposes that the total traffic routed on each ring does not exceed the capacity $B$. In other words, for a given ring $i$, it forces the total traffic demands of all the customers connected to this ring, to be lower or equal to the bandwidth.
Constraint (\ref{eq:consSRAP2}) forces the load of federal ring to be less than or equal to $B$. To do so, it computes the sum of the traffic demands between all the pairs of customers connected to different rings.

\begin{figure}[h]
 \begin{center}
  \includegraphics[scale=0.8]{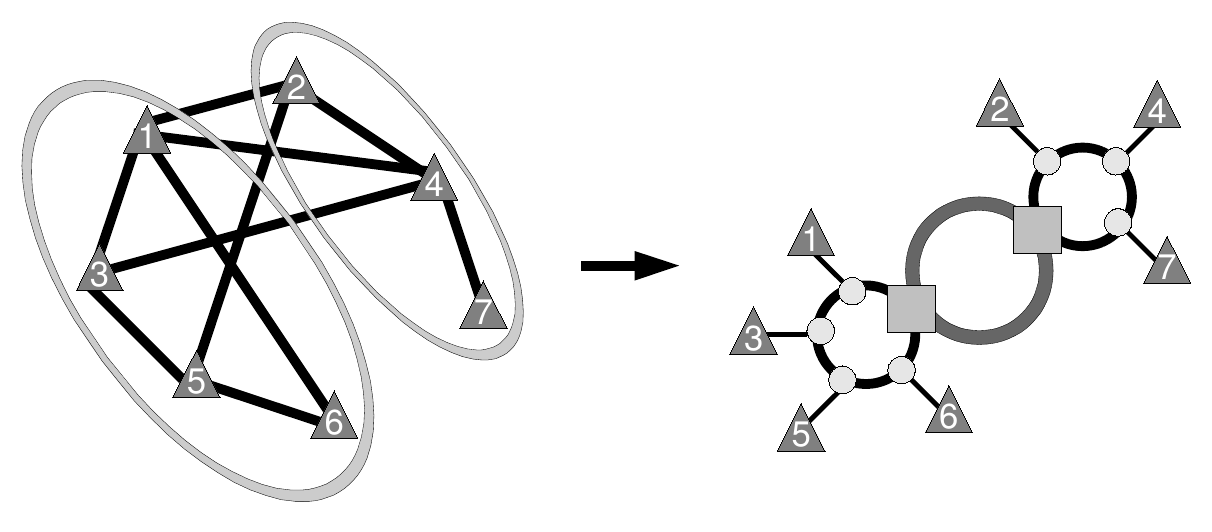}
  \caption{Relation between the node partitioning and the network topology. \label{fig:nptop}}
 \end{center}
\end{figure}

Figure \ref{fig:nptop} illustrates the relation between the node partitioning model and the first topology SRAP. We can see that, because the nodes $1$, $3$, $5$ and $6$ are in the same partition, they are connected to the same ring. Similarly, the nodes $2$, $4$ and $7$ are on the same ring. 

For this problem we can easily compute a lower bound $k_{lb}$ introduced in \cite{GLO2003}. In fact, we want to know the minimum number of partitions needed to route all the traffic. Reasonning on the total traffic amount, if we sum all the traffic demands of the graph and divide it by the bandwidth $B$, we trivially obtain a minmum for the number of rings, that is, a lower bound of the number of partitions. Moreover, we cannot have fractional part of partition, that is why we take the upper round of this fraction.
$$k_{lb} = \left\lceil \frac{\displaystyle{\sum_{u = 1}^{n-1} \sum_{v = u + 1}^n d_{uv}}}{B} \right\rceil$$

\subsubsection{Second topology (IDP)} 
\begin{figure}[h]
 \begin{center}
  \includegraphics[scale=0.6]{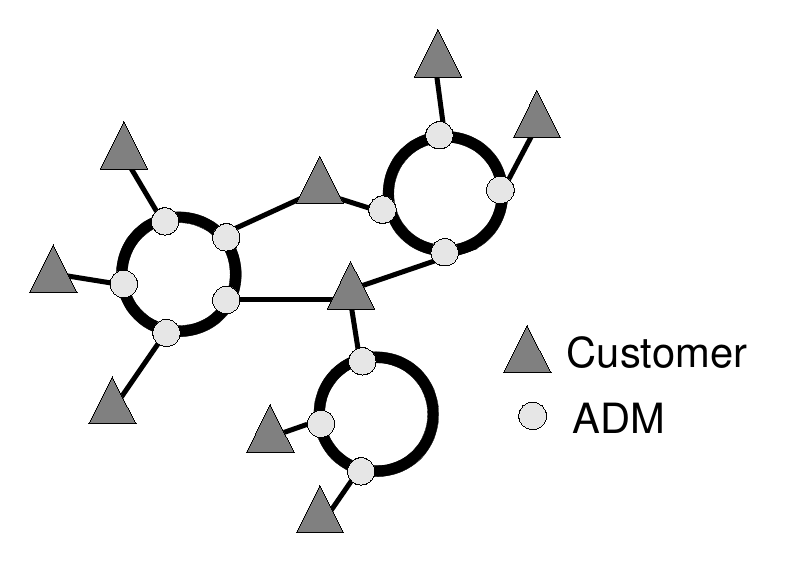}
  \caption{A SONET network without DXC. \label{fig:top2}}
 \end{center}
\end{figure}

In the second topology, customers can be connected to more than one ring. If two customers want to communicate, they have to be connected to the same ring. In this case, the DXC are no longer needed and neither is the federal ring. However there are more ADM used than in the first topology. In this case, the most expensive component is the ADM although its price has significantly dropped over the past few years. It is important, in this topology, to have the smallest numbers of ADMs. This problem is called \textit{Intra-ring Synchronous Optical Network Design Problem} (IDP). The figure \ref{fig:top2} illustrates this topology.

\paragraph{Model} Contrarily to the SRAP problem, there is no need to assign each customer to a particular ring because customers can be connected to several rings. Here the model is based on a partition of the edges of the graph, where a subset of the partition corresponds to a ring.

Formally, let $E_1, E_2, \dots, E_k$ be a partitioning of $E$ in $k$ subsets and $Nodes(E_i)$ be the set of endpoint nodes of the edges in $E_i$. Each subset of the partition corresponds to a ring, in other words, each customer in $Nodes(E_i)$ is linked to the $i$-th ring. In the corresponding IDP network, there are $\displaystyle{\sum_{i=1}^k|Nodes(E_i)|}$ ADM and no DXC.

Hence, minimizing the number of ADMs is equivalent to minimizing $\displaystyle{\sum_{i=1}^k|Nodes(E_i)|}$ subject to,
\begin{equation}
  \sum_{(u,v) \in E_i}d_{uv} \leq B, \qquad \forall i = 1, \dots, k \label{eq:consIDP1}
\end{equation}
Constraint (\ref{eq:consIDP1}) imposes that the traffic in each ring does not exceed the capacity $B$.

\begin{figure}[h]
 \begin{center}
  \includegraphics[scale=0.8]{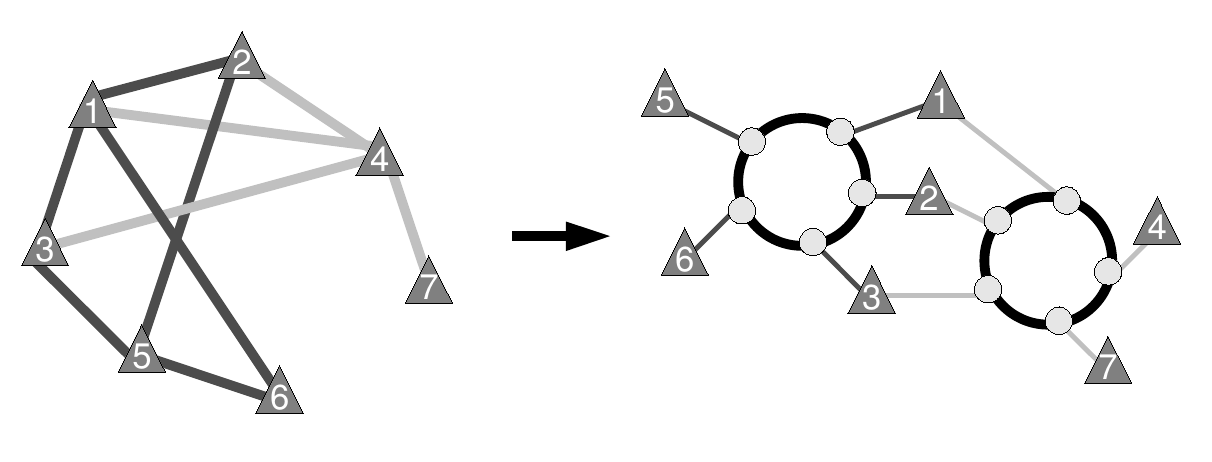}
  \caption{Relation between the edge partitioning and the network topology. \label{fig:eptop}}
 \end{center}
\end{figure}

Figure \ref{fig:eptop} shows the relation between the edge partitioning and the second topology. If all the edges of a node are in the same partition, this node will only be connected to a ring. We can see, for example, the node $4$ has all its edges in the same partition, because of that, the node $4$ is connected to only one ring. On the opposite, the edges of the node $2$ are in two different partitions, so it is connected to two rings.

\paragraph{}
The SRAP problem can be seen as a node partitioning problem, whereas IDP, as an edge partitioning problem for the graph described above, subject to capacity constraints. These graph partitioning problems have been introduced in \cite{GLO2003} and \cite{GHLO2003}.

Both of these problems are $\mathcal{NP}$-hard (see O. Goldschmidt, A. Laugier and E. Olinick in 2003, \cite{GLO2003}, and O. Goldschmidt, D. Hochbaum, A. Levin and E. Olinick in 2003, \cite{GHLO2003} for details). The principal constraint, the load constraint, is similar to a capacity constraint, yet different: a capacity constraint holds on the variables \textit{in } the sum, while the load constraint holds on the variables \textit{below } the sum. The question is how to choose the $d_{uv}$ (which are data) that count for the load.

\subsection{Brief introduction of Local Search \label{sec:LS}}
In order to efficiently and quickly solve these two combinatorial optimization problems, we decided to use Local Search instead of an exact algorithm. Indeed, it permits to search in a efficiently way among all the candidate solutions, by performing steps from a solution to another.

\paragraph{Principles}
Local search is a metaheuristic based on iterative improvement of an objective function. It has been proved very efficient on many combinatorial optimization problems like the Maximum Clique Problem (L. Cavique, C. Rego and I. Themido in 2001 in \cite{CRT2001}), or the Graph Coloring Problem (J.P. Hansen and J.K. Hao in 2002 in \cite{HH2002}). It can be used on problems which formulated either as mere optimization problems, or as constrained optimization problems where the goal is to optimize an objective function while respecting some constraints. Local search algorithms perform local moves in the space of candidate solutions, called the search space, trying to improve the objective function, until a solution deemed optimal is found or a time bound is reached. Defining the neighborhood graph and the method to explore it are two of the key ingredients of local search algorithms.

The approach for solving combinatorial optimization problems with local search is very different from the systematic tree search of constraint and integer programming. Local search belongs to the family of metaheuristic algorithms, which are incomplete by nature and cannot prove optimality. However on many problems, it will isolate a optimal or high-quality solution in a very short time: local search sacrifices optimality guarantees to performance. In our case, we can compute the lower bound to either prove that the obtained solution is optimum, or estimate its optimality, hence local search is well suited.

\paragraph{Basic algorithm}
A local search algorithm starts from a candidate solution and then iteratively moves to a neighboring solution. This is only possible if a neighborhood relation is defined on the search space. Typically, for every candidate solution, we define a subset of the search space to be the neighborhood. Moves are performed from neighbors to neighbors, hence the name local search. The basic principle is to choose among the neighbors the one with the best value for the objective function. The problem is then that the algorithm will be stuck in local optima. Metaheuristics, such as Tabu Search, are added to avoid this. In Tabu Search, the last $t$ visited configurations are left out of the search ($t$ being a parameter of the algorithm): this ensures that the algorithm can escape local optima, at least at order $t$. A pseudo-code is given on figure \ref{ag:tb}.

\begin{algorithm}
  \decmargin{\parindent}
  \SetNoline
  Choose or construct an initial solution $S_0$ \;
  $S \leftarrow S_0$ \tcc*{$S$ is the current solution}
  $S^* \leftarrow S_0$ \tcc*{$S^*$ is the best solution so far}
  $bestValue \leftarrow objValue(S_0)$ \tcc*{$bestValue$ is the evaluation of $S^*$}
  $T \leftarrow \emptyset$ \tcc*{$T$ is the Tabu list}
  \While{Terminaison criterion not satisfied}{
    $N(S) \leftarrow $ all the neighboring solutions of $S$ \tcc*{Neighborhood exploration}
    $S \leftarrow $ a solution in $N(S) $ minimizing the objective \;
    \If(\tcc*[f]{The solution found is better than $S^*$}){$objValue(S) < bestValue$}{
      $S^* \leftarrow S$ \;
      $bestValue \leftarrow objValue(S)$ \;
    }
    Record tabu for the current move in T (delete oldest entry if necessary) \;
  }
  \caption{Tabu Search \label{ag:tb}}
\end{algorithm}

Termination of local search can be based on a time bound. Another common choice is to terminate when the best solution found by the algorithm has not been improved in a given number of iterations. Local search algorithms are typically incomplete algorithms, as the search may stop even if the best solution found by the algorithm is not optimal. This can happen even if termination is due to the impossibility of improving the solution, as the optimal solution can lie far from the neighborhood of the solutions crossed by the algorithms.

\subsection{\comet}
\comet is an object-oriented language created by Pascal Van Hentenryck and Laurent Michel. It has a constraint-based architecture that makes it easy to use when implementing local search algorithms, and more important, constraint-based local search algorithms (see \cite{VHM2005} for details).

Moreover, it has a rich modeling language, including invariants, and a rich constraint language featuring numerical, logical and combinatorial constraints. Constraints and objective functions are differentiable objects maintaining the properties used to direct the graph exploration. The constraints maintain their violations and the objectives their evaluation. One of its most important particularity, is that differentiable objects can be queried to determine incrementally the impact of local moves on their properties.

As we can see on the constraint (\ref{eq:consSRAP1}), the sum are on datas ($d_{uv}$) and are determined by the variables ($u \in V_i, v \in V, v \neq u$). We will rely on \comet's built-in invariants to define a constraint to represent the load.

\section{Related work \label{sec:rwork}}
These two problems have been well studied. It has been proven that they are both $\mathcal{NP}$-hard (\cite{GLO2003}, \cite{GHLO2003}).

\subsection{Greedy algorithms for SRAP}
In \cite{GLO2003} the SRAP problem is considered. They propose three greedy algorithms with different heuristics, the \textit{edge-based}, the \textit{cut-based} and the \textit{node-based}. The first two algorithms start by assigning each node to a different ring. At each iteration they reduce the number of rings by merging two rings $V_i$ and $V_j$ if $V_i \cup V_j$ is a feasible ring for the capacity constraint. In the edge-based heuristic, the two rings with the maximum weight edge are merged. While in the cut-based heuristic, the two rings with the maximum total weight of the edges with one endpoint in each of them, are merged. Algorithm \ref{ag:eb} shows the pseudo code for the edge-based heuristic.

\begin{algorithm}
  \decmargin{\parindent}
  \SetNoline
  $F \leftarrow E$ \tcc*{Initialize the set of edges that have not been used yet}
  $\forall v \in V~ring(v) \leftarrow v$ \tcc*{Assign each node to a different ring}
  \While(\tcc*[f]{There is still some edges that have not been used}){$F \neq \emptyset$}{
    Choose a maximum capacity edge $(u, v) \in F$ \;
    $i \leftarrow ring(u), j \leftarrow ring(v)$ \;
    \eIf(\tcc*[f]{Merging the rings gives a feasible ring}){ $V_i \cup V_j$ ~is a feasible ring}{
      $\forall v \in V_j~ring(v) \leftarrow i$ \;
      $F \leftarrow F \backslash \{(x, y) | ring(x) = i, ring(y) = j\}$ \;
    }{
      $F \leftarrow F \backslash \{(u, v)\}$ \;
    }
  }
  \caption{Edge-Based Heuristic \label{ag:eb}}
\end{algorithm}

Given a value $k$, the node-based heuristic, starts by randomly assigning a node to each of the $k$ rings. At each iteration it first chooses the ring $V_i$ with the largest unused capacity, then the unassigned node $u$  with the largest traffic with the nodes in $V_i$. Finally it adds $u$ to the ring $V_i$ disregarding the capacity constraint. The pseudo-code for this heuristic is shown on algorithm \ref{ag:nb}. 
\\
The node-based heuristic is run ten times. At each run, if a feasible solution is found, the corresponding value for $k$ is kept and the next run takes $k-1$ as an input. The idea behind this is to try and improve the objective at each run.

\begin{algorithm}
  \decmargin{\parindent}
  \dontprintsemicolon
  \SetNoline
  $U \leftarrow V$ \tcc*{Initialize the set of nodes that have not been used yet}
  \For(\tcc*[f]{Assign $k$ random nodes to the $k$ partitions}){$i = 1~\KwTo~k$}{
    Choose~$u \in U, V_i \leftarrow u, U \leftarrow U \backslash \{u\}$\\
  }
  \While(\tcc*[f]{There are some unused nodes}){$U \neq \emptyset$}{
    Choose a minimum capacity ring~$V_i$\\
    Choose~$u \in U$~to maximize $\displaystyle{\sum_{\{v \in V_i\}}d_{uv}}$\\
    $ring(u) \leftarrow V_i, U \leftarrow U \backslash \{u\}$ \tcc*{Assign $u$ to $V_i$}
  }
   \caption{Node-Based Heuristic \label{ag:nb}}
\end{algorithm}

To test these heuristics, the authors have randomly generated 160 instances\footnote{These instances are available at \url{www.seas.smu.edu/~olinick/srap/GRAPHS.tar.}}. The edge-based, and the cut-based are run first. If they have found a feasible solution and obtained a value for $k$, the node-based is then run with the smallest value obtained for $k$ as input. If they have not, the node-based heuristic has for input a random value from the range $[k_{lb}, |V|]$ where $k_{lb}$ is the \textit{lower bound}, described previously.

\subsection{MIP and Branch and Cut for IDP}
A special case of the IDP problem where all the edges have the same weight, is studied in \cite{GHLO2003}. This special case is called the \textit{K-Edge-Partitioning} problem. Given a simple undirected graph $G = (V, E)$ and a value $k < |E|$, we want to find a partitioning of $E$, $\{E_1, E_2, \dots\, E_l\}$ such that $\forall i \in \{1, \dots, l\}, |E_i| \leq k$. The authors present two linear-time-approximation algorithms with fixed performance guarantee.

Y. Lee, H. Sherali, J. Han and S. Kim in 2000 (\cite{LSHK2000}), have studied the IDP problem with an additional constraint such that for each ring $i$,  $|Nodes(E_i)| \leq R$.  The authors present a mixed-integer programming model for the problem, and develop a branch-and-cut algorithm. They also introduce a heuristic to generate an initial feasible solution, and another one to improve the initial solution. To initialize a ring, the heuristic first, adds the node $u$ with the maximum graph degree, with respect to unassigned edges, and then adds to the partition the edge $[u, v]$ such that the graph degree of $v$ is maximum. It iteratively increases the partition by choosing a node such that the total traffic does not exceed the limit $B$. A set of 40 instances is generated to test these heuristics and the branch-and-cut.

\subsection{Local Search for SRAP and IDP}
More recently, in \cite{AD2005}, these two problems have been studied. The authors have developed different metaheuristic algorithms, all based on the Tabu Search. The metaheuristics are the \textit{Basic Tabu Search} (BTS), two versions of the \textit{Path Relinking} (PR1, PR2), the \textit{eXploring Tabu Search} (XTS), the \textit{Scatter Search} (SS), and the \textit{Diversification by Multiple Neighborhoods} (DMN). These local search algorithms are detailed further.

Previously, we saw that with local search it is necessary to define a neighborhood to choose the next solution. The authors of \cite{AD2005} use the same for all of their metaheuristics. It tries to assign an item $x$ from a partition, $P_1$, to another partition, $P_2$. The authors also consider the neighborhood obtained by swapping two items, $x$ and $y$, from two different partitions, $P_1$ and $P_2$. But instead of trying all the pairs of items, it will only try to swap the two items if the resulting solution of the assignment of $x$ to the partition $P_2$ is unfeasible.

In order to compute a starting solution for the IDP problem, the authors describe four different heuristics.
The first heuristic introduced in \cite{AD2005} ordered the edges by decreasing weight, at each iteration it tries to assign the edge with the biggest weight which is not already assigned, to the ring with the smallest residual capacity regarding to capacity constraint. If no assignment is possible, the current edge is assigned to a new ring.
The second one, sorts the edges by increasing weight, and tries to assign the current edge to the current ring if the capacity constraint is respected, otherwise the ring is no longer considered and a new ring is initialized with the current edge.

The two other methods described in \cite{AD2005} are based on the idea that to save ADMs a good solution should have very dense rings. They are both greedy and rely on a clique algorithm. In graph theory, a clique in an undirected graph $G = (V, E)$ is a subset of the vertex set $C \subseteq V$, such that for every two vertices in $C$, there exists an edge connecting the two. Finding a clique is not that easy, a way to do it is to use an "Union-Find" strategie, \textit{Find} two clique $A$ and $B$ such that each node in $A$ is adjacent to each node in $B$ then merge the two cliques (\textit{Union}). The associated heuristic starts by considering each node to be a clique of size one, and to merge two cliques into a larger clique until there are no more possible merges. 

In the third method, \textit{Clique-BF}, it iteratively selects a clique of unassigned edges with the total traffic less or equal to $B$. Then assigns it to the ring that minimizes the residual capacity and, if possible, preserves the feasibility. If both of them are impossible it places it to a new ring. Algorithm \ref{ag:cbf} shows the pseudo code associated to this heuristic.
The last algorithm, \textit{Cycle-BF}, is like the previous method, but instead of looking for a clique at each iteration it try to find a cycle with as many cords as possible.

\begin{algorithm}
  \decmargin{\parindent}
  \SetNoline
  $U \leftarrow E$ \;
  $r \leftarrow 0$ \;
  \While{$U \neq \emptyset$}{
    Heuristicaly find a clique~$C \subset U$~such that $weight(C) \leq B$ \; \tcc{Search a ring such that the weight of the ring plus the weight of the clique does not exceed $B$ and is the biggest possible}
    $j \leftarrow min\{B - weight(E_i) - weight(C): i \in \{1, \dots, k\}, B - weight(E_i) - weight(C) \geq 0\}$ \;
    \If{$j =$~null}{$r++$ \;$j \leftarrow r$ \;}
    $E_j \leftarrow E_j \cup C$ \;
    $U \leftarrow U \backslash C$ \;
  }
  \caption{Clique-BF \label{ag:cbf}}
\end{algorithm}

They also introduce four objective functions, one of which depends on the current and the next status of the search. Let $z_0$ be the basic objective function counting the number of rings of a solution for SRAP, and the total number of ADMs for IDP, and let $BN$ be the highest load of a ring in the current solution.

\begin{eqnarray}
  z_1 & = & z_0 + max\{0, BN - B\},\nonumber \\
  z_2 & = & z_1 + \left\{\begin{array}{ll}
                           \alpha\cdot RingLoad(r) \quad & \mbox{if the last move has created a new ring }r,\\
                           0 & \mbox{otherwise}
                         \end{array} \right.\nonumber \\
  z_3 & = & z_0 \cdot B + BN \nonumber \\
  z_4 & = & \left\{\begin{array}{ll}
                     z_{4a} = z_0 \cdot B + BN (= z_3) & \mbox{(a): from feasible to feasible}\\
                     z_{4b} = (z_0 + 1)BN & \mbox{(b): from feasible to unfeasible}\\
                     z_{4c} = z_0B & \mbox{(c): from unfeasible to feasible}\\
                     z_{4d} = \beta z_0BN & \mbox{(d): from unfeasible to unfeasible}
                   \end{array} \right.\nonumber
\end{eqnarray}
with $\alpha \geq 1$ and $\beta \geq 2$, two fixed parameters, and where $RingLoad(r)$ is the load of the ring $r$.

The first function $z_1$ minimizes the basic function $z_0$. As $BN > B$, it also penalizes the unfeasible solutions, by taking into account only one ring, the one with the highest overload. In addition to the penalty for the unfeasible solutions, $z_2$ penalizes the moves that increase the number of rings. Function $z_3$ encourages solutions with small $z_0$, while among all the solutions with the same value of $z_0$, it prefers the ones in which the rings have the same loads. The last objective function $z_4$ is an adapting technique that modifies the evaluation according to the status of the search. It is a \textit{variable objective function} having different expressions for different transitions from the current status to the next one.

\section{Our work \label{sec:work}}
In this section we present the different tools needed to implement the Constraints Based Local Search algorithms for SRAP and IDP. First we introduce the starting solution, then the neighborhoods and the objective functions. Finally we present the different local search algorithms.

\subsection{Starting solution}

Most of the times, local search starts from an random initial solution. However we have tested other possibilities and two other options proved to be more efficient.

The best initializing method assigned all the items, nodes for SRAP or edges for IDP, to the same partition. This solution is certainly unfeasible as all the traffic is on only one ring. This biases the search towards solutions with a minimum value for the cost and a very bad value for the capacity constraints' violations. Astonishingly this is the one that gave us the best results on large instances.

We had good confidence in another one which first computes the lower bound $k_{lb}$ (described in section \ref{sec:rwork}) and randomly assigns all the items to exactly $k_{lb}$ partitions. The idea was to let the Local Search reduce the number of violations. This starting solution was good on small instances and not so good on large ones.
It was the same with a random solution, which corresponds, for these problems, to a solution where all the items are randomly assigned to a partition.

\subsection{Neighborhoods \label{sec:neigh}}
In a generic partitioning problem there are usually two basic neighborhoods. From a given solution, we can move an object from a subset to another subset or swap two objects assigned to two different subsets.
For SRAP a neighboring solution is produced by moving a node from a ring to another (including a new one) or by swapping two nodes assigned to two different rings.
The same kind of neighborhood can be used for IDP: moving an edge from a ring to another or swapping two edges.

In some cases it is more efficient to restrain the neighborhood to the feasible space.
We have tested different variants of the basic neighborhood applying this idea, by choosing the worst partition (wrt. the capacity constraint) and even by assigning it to the partition with the lowest load. Anyway it appears to be less efficient than the basic one. As will be seen later it seems that on these problems it is necessary to keep the search as broad as possible.

\subsection{Objective function \label{sec:of}}
We have compared the four objective functions described in \cite{AD2005} (see Section \ref{sec:rwork}) to a new one we have defined: $z_5$. $$z_5 = z_0 + \sum_{p~\in~partitions}violations(p)$$ where 

$partitions$ are all the rings (in the case of the SRAP problem the \textit{federal ring} is also included), 
$$violations(p) =
  \left\{
    \begin{array}{ll}
      capacity(p) - B \qquad\qquad  & \mbox{if the load of } p \mbox{ exceed } B\qquad\qquad\qquad\qquad\quad\\
      0 & \mbox{otherwise.}\\
    \end{array}
  \right.
$$

This objective function minimizes the basic function $z_0$ and penalizes the unfeasible solutions, but contrarily to the previous objectives, this penalty is based on all the constraints. We consider that every constraint is violated by a certain amount (its current load minus $B$). By summing all the violations of the current solution, we obtain the total violation for all the constraints, and we can precisely say how far we are from a feasible one. If the current solution is feasible, $\displaystyle{\sum_{p~\in~partitions}violations(p) = 0}$.

This objective has also the nice property that it is merely local, depending only on the current solution and not on the other moves. Notice that a feasible solution with 4 rings will be preferred to an unfeasible solution with 3 rings, as $z_0$ is much smaller than the load of a ring.

\subsection{Local Search \label{sec:ls}}

We have proposed a new algorithm called DMN2 which proved to be efficient on both problems. It is a variant of the Diversification by Multiple Neighborhood (DMN) proposed in \cite{AD2005}. DMN is based on Tabu Search, and adds a mechanism to perform diversification when the search is going round and round without improving the objective (eventhough it is not a local minimum). This replaces the classical random restart steps. We refine this particular mechanism by proposing several ways of escaping such areas.

More precisely, on our problems, after a series of consecutive non improving iterations, the DMN algorithm empties a partition by moving all its items to another partition, disregarding the capacity constraint and locally minimizing the objective function. There is a particular case for our function $z_5$, because it integrates the capacity constraints. In this case, the "$z_5$" version of DMN we have implemented moves the items to another partition minimizing $z_5$.
The results in \cite{AD2005} show a general trend on SRAP and IDP: the more diversification is performed, the better are the results.
Following this idea, we propose different ways of perfoming the DMN step, which gives our algorithm DMN2. In DMN2, when the search needs to be diversified, it randomly chooses among three diversification methods ($d_1, d_2, d_3$). The first method, $d_1$, is the diversification used in DMN. The second one, $d_2$, generates a random solution, in the same way as a classic random restart. Finally, $d_3$ randomly chooses a number $m$ in the range [1, $k$], where $k$ is the number of rings, and applies $m$ random moves.

\begin{algorithm}
  \decmargin{\parindent}
  \SetNoline
  Construct the initial solution $S_0$ \;
  $S \leftarrow S_0$ \tcc*{$S$ is the current solution}
  $S^* \leftarrow S_0$ \tcc*{$S^*$ is the best solution so far}
  $bestValue \leftarrow objValue(S_0)$ \tcc*{$bestValue$ is the evaluation of $S^*$}
  $T \leftarrow \emptyset$ \tcc*{$T$ is the Tabu list}
  $nonImprovingStep \leftarrow 0$ \; \tcc{$nonImprovingStep$ is the number of consecutive non improving iterations}
  \While{$bestValue > lower bound$ and we still have time}{
    $N(S) \leftarrow $ all the neighboring solution of $S$ \;
    $S \leftarrow $ the solution in $N(S) $ minimizing the objective \;
    \eIf(\tcc*[f]{The solution found is better than $S^*$}){$objValue(S) < bestValue$}{
      $S^* \leftarrow S$ \;
      $bestValue \leftarrow objValue(S)$ \;
      $nonImprovingStep \leftarrow 0$ \;
    }(\tcc*[f]{The iteration did not improve $S^*$}){
      $nonImprovingStep++$ \;
    }
    Record tabu for the current move in T (delete oldest entry if necessary) \;
    \If{$nonImprovingStep = maxNonImprovingStep$}{
      \tcc{Stuck in a non improving area: time for diversification} $nonImprovingStep \leftarrow 0$ \;
       \tcc{Uniformly chooses the diversification to apply} $d \leftarrow UniformRandom(1,3)$ \;
      \Switch{The value of d}{
        \uCase(\tcc*[f]{Empties a partition}){1}{
          Randomly choose a partition $p$ \;
          \ForEach{item $i$ on the partition $p$}{
            Choose the partition $p'$ such that the assignment of $i$ to $p'$ minimizes the objective function and $p' \neq p$ \;
            Assign $i$ to the partition $p'$ \;
          }
        }
        \uCase(\tcc*[f]{Random Restarts}){2}{
          \ForEach{item $i$}{
            Randomly choose a partition $p$ \;
            Assign $i$ to the partition $p$ \;
          }
        }
        \Case(\tcc*[f]{Applies $m$ random moves}){3}{
          $m \leftarrow UniformRandom(1,numberOfRings)$ \;
          Apply $m$ moves \;
        }
      }
    }
  }
  \caption{DMN2 \label{ag:dmn2}}
\end{algorithm}

\paragraph{}
In the end, our general algorithm starts with a solution where all the items are in the same partition. Then it applies one of the local search algorithms described before. If the solution returned by the local search is feasible but with the objective value greater than the lower bound $k_{lb}$, it empties one partition by randomly assigning all its items to another. Then run once again the local search until it founds a solution with the objective value equals to $k_{lb}$ or until the time limit is exceeded. 

\section{Results \label{sec:results}}
The objective functions and the metaheuristics, respectively described in Section \ref{sec:of} and Section \ref{sec:ls}, have been coded in \comet and tested on Intel based, dual-core, dual processor, Dell Poweredge 1855 blade server, running under Linux. The instances used are from the litterature.

\subsection{Benchmark}
To test the algorithms, we used two sets of instances.
The first one has been introduced in \cite{GLO2003}. They have generated 80 \textit{geometric} instances, based on the fact that customers tend to communicate more with their close neighbors, and 80 \textit{random} instances. These subsets have both 40 \textit{low-demand} instances, with a ring capacity $B = $ 155 Mbs, and 40 \textit{high-demand} instances, where $B = $ 622 Mbs. The traffic demand between two customers, $u$ and $v$, is determined by a discrete uniform random variable corresponding to the number of T1 lines required for the anticipated volume of traffic between $u$ and $v$. A T1 line has an approximate capacity of 1.5 Mbs. The number of T1 lines is randomly picked in the interval $[3, 7]$, for low-demand cases, while it is selected from the range $[11, 17]$, for the high-demand cases. The generated graphs have $ |V| \in \{ 15, 25, 30, 50 \} $. In the 160 instances, generated by O. Goldschmidt, A. Laugier and E. Olinick in 2003, 42 have been proven to be unfeasible by R. Aringhieri and M. Dell'Amico using CPLEX 8.0 (see \cite{AD2005}).

The second set of instances has been presented in \cite{LSHK2000}. They have generated 40 instances with a ring capacity $B = 48 \times $ T1 lines and the number of T1 lines required for the traffic between two customers has been chosen in the interval $[1, 30]$. The considered graphs have $ |V| \in \{ 15, 20, 25 \} $ and $ |E| = \{ 30, 35 \}$. Most of the instances in this set are unfeasible.

Note that all the instances can be feasible for the IDP problem, we always could assign each demand to a different partition.

\subsection{Computational Results}

We now describe the results obtained for SRAP and IDP on the above two benchmark sets, by the algorithms Basic Tabu Search (BTS), Path Relinking (PR1, PR2), eXploring Tabu Search (XTS), Scatter Search (SS), Diversification by Multiple Neighborhoods (DMN, DMN2) (see Section \ref{sec:ls} for details). For each algorithm we consider the five objective functions of Section \ref{sec:of}, but for the SS we use the three functions described in Section \ref{sec:ls}.

We gave a time limit of $5$ minutes to each run of an algorithm. However we observed that the average time to find the best solution is less than $1$ minute. Obviously, the algorithm terminates if the current best solution found is equal to the lower bound $k_{lb}$. In case the lower bound is not reached, we define as a \textit{high-quality solution} a solution for which the evaluation of the objective is equal to $k_{lb} + 1$. Remind that objective functions $z_2$ and $z_3$ cannot be applied with the Scatter Search.

\begin{figure}[h]
 \begin{center}
    \includegraphics[scale=0.6]{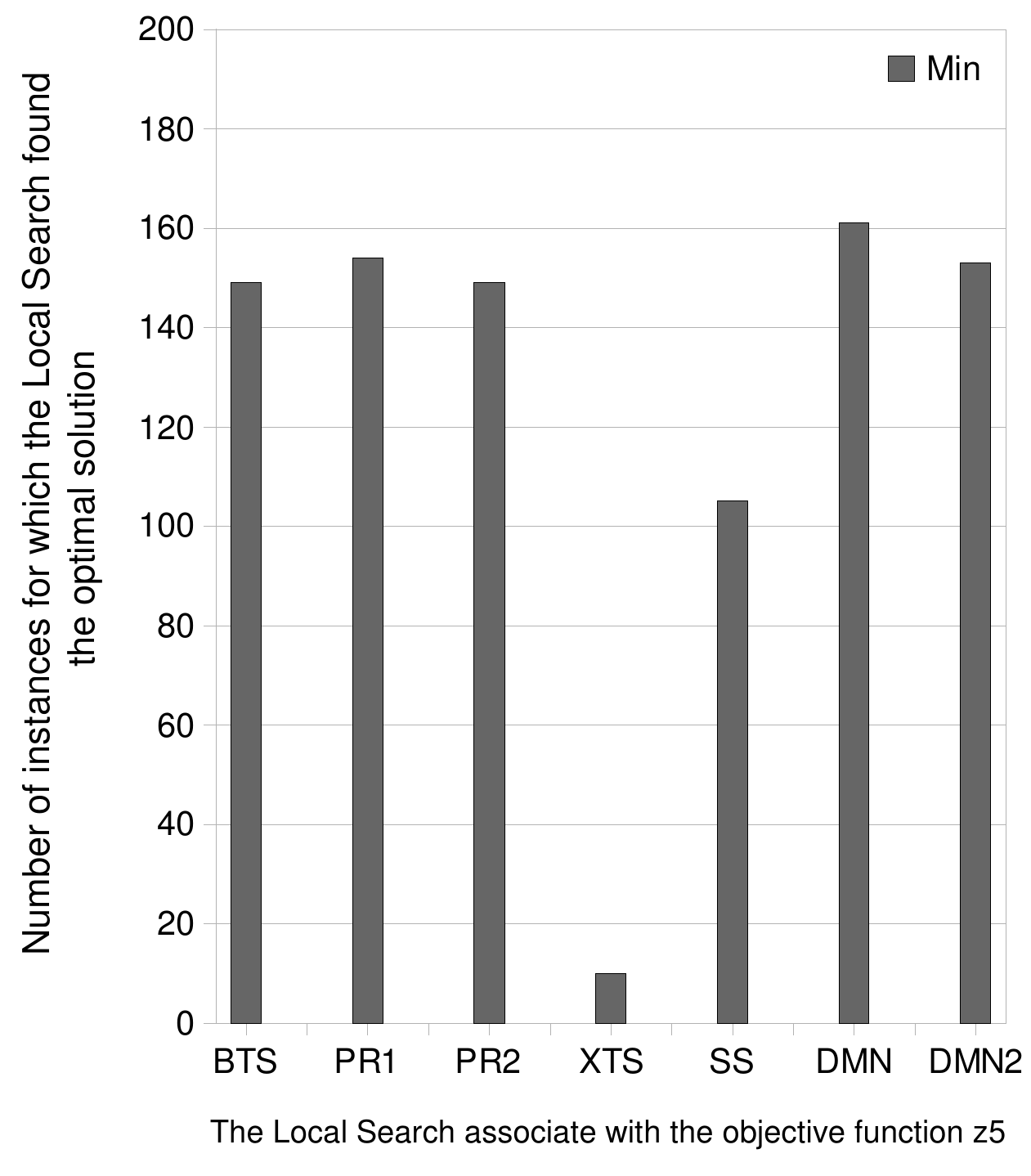}
    \caption{Results for IDP. \label{fig:resIdp}}
 \end{center}
\end{figure}

The figure \ref{fig:resIdp} only shows for each algorithm the number of optimal solutions found with the objective function $z_5$. With the other objectives, the number of optimal solutions found is zero, that is why we did not show them on the diagram. However the other objectives found good solutions. Our conclusion is that maybe the other functions do not enough discriminate the different solutions.
For this problem, we can see that the eXploring Tabu Search does not give good results. This can be due to a too early "backtracking". After a fixed number of consecutive non improving iterations the search goes back in a previous configuration and applies the second best move. In the case of the IDP problem, it could take much more iterations to improve the value of the objective function than for the SRAP problem. Indeed, the value of the objective function depends on the number of partitions in which a customer belongs, while an iteration moves only one edge ; and to reduce its value by only one it could need to move several edges.

\begin{figure}[h]
 \begin{center}
    \includegraphics[scale=0.7]{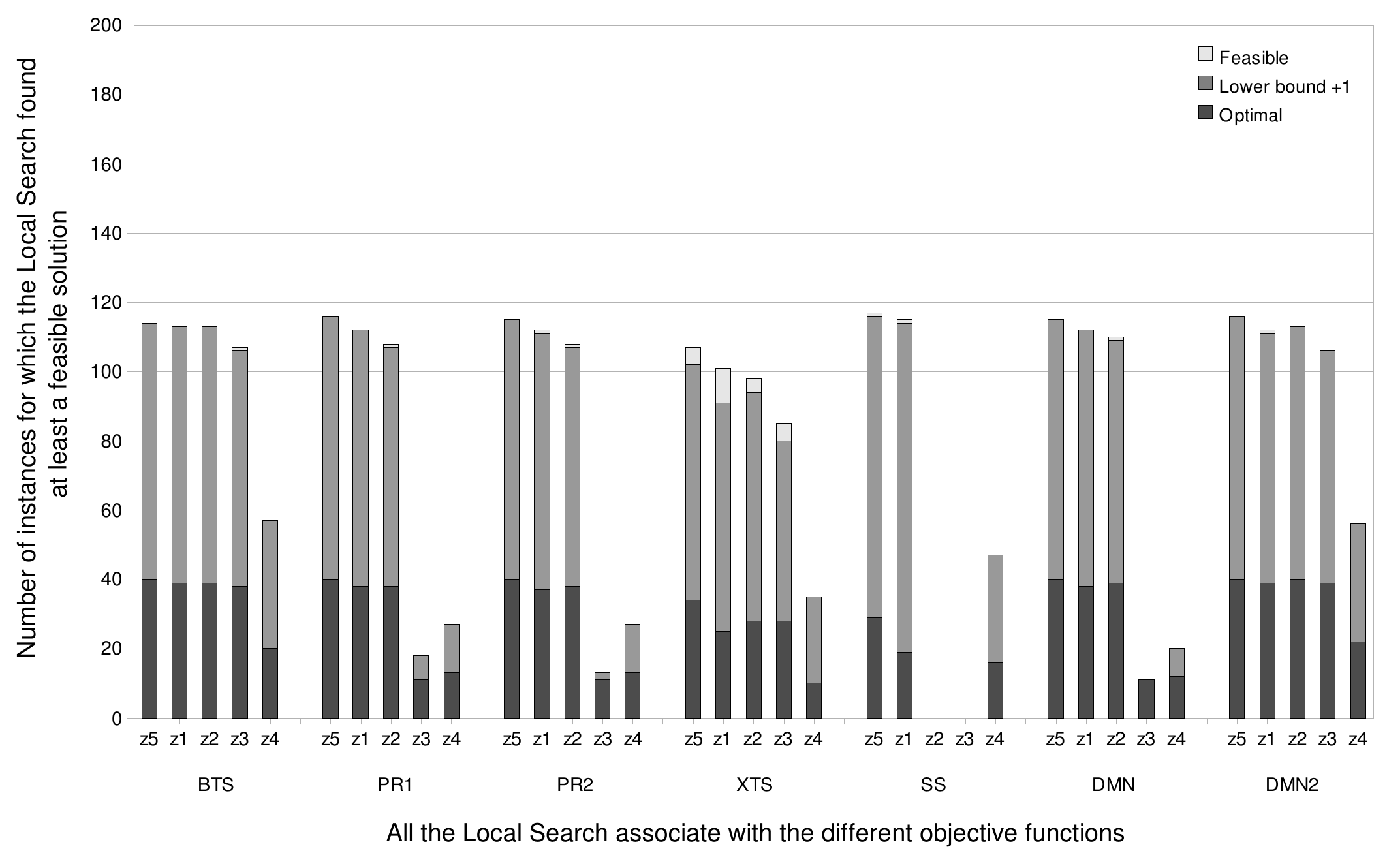}
    \caption{Results for SRAP. \label{fig:resSrap}}
 \end{center}
\end{figure}

Figure \ref{fig:resSrap} shows for each algorithm and each objective function, the number of instances for which the search has found an optimal solution, i.e. a solution with $k_{lb}$ partitions (in dark gray on the diagram) ; the number of those for which the best feasible solution found has $k_{lb} + 1$ partitions (in gray) ; and, in light gray, the number of instances for which it has found a feasible solution with more than $k_{lb} + 1$ partitions. From the objective functions perspective, we can see that $z_4$, supposed to be the most improving one, is not that good in the \comet implementation. However the one we add, $z_5$, is always better than the other ones.

Against all odds, the Basic Tabu Search on all the objective functions, is as good as the other search algorithms. Still on the local search algorithms, we can see that the second version of the Diversification by Multiple Neighborhoods, is much better than the first one with the objectives $z_3$ and $z_4$.

For the details of our results see the report \cite{P2009}.

\section{Conclusion \label{sec:conc}}
The purpose of this work was to reproduce with \comet the results obtained, for the SONET Design Problems, by  R. Aringhieri and M. Dell’Amico in 2005 in ANSI C (see \cite{AD2005} for details).

We have implemented in \comet the algorithms and the objective functions  described in this paper. We found relevant to add a variant of one of their local search algorithm and a new objective function. Unfortunately, we cannot exactly compare our results to theirs because the set of $230$ instances they have generated is not available. 
However, for the IDP problem, we obtained better results for $15$ instances over the $160$ compared, and similar results for the other instances. Unfortunately we did not found their results for the SRAP problem. Still for the problem SRAP, compare to the results obtained by O. Goldschmidt, A. Laugier and E. Olinick in 2003, \cite{GLO2003} we obtained better results, we have more instances for wich the algorithm reach the lower bound and less unfeasible instances.
It would be interesting to have all the instances and the results to fully compare our results.

In the end we can exhibit two main observations. Firstly, for these two problems, the more an algorithm uses diversification the better it is. Actually, we have tried different intensification methods for the local search algorithms but none of them improved the results, worst, they gave us pretty bad results.

Secondly, based on our results, we can say that our objective function implemented in \comet finds more good solutions than the other ones.
It is a constraint-based objective function taking into account the violation of every constraint. Hence it has the asset of being both more generic and precise than the dedicated functions, with better results.

%
%
\bibliographystyle{eptcs} 

\begin{thebibliography}{1}

\bibitem {VHM2005}
Van Hentenryck, Pascal and Michel, Laurent (2005):
\newblock \emph{Constraint-Based Local Search}.
The MIT Press

\bibitem {AD2005}
Aringhieri, Roberto, Dell'Amico, Mauro (2005):
\newblock \emph{Comparing Metaheuristic Algorithms for Sonet Network Design Problems}.
\newblock {\sl Journal of Heuristics} 11, pp. 35--57

\bibitem {DT1998}
Dell'Amico, Mauro, Trubian, Marco (1998):
\newblock \emph{Solution of Large Weighted Equicut Problems}.
\newblock {\sl European J. Oper. Res} 106(2/3), pp. 500--521

\bibitem {GL1997}
Glover, Fred, Laguna, Manuel (1997):
\newblock \emph{Tabu Search}.
Boston. Kluwer Academic Publishers

\bibitem {G1997}
Glover, Fred (1997):
\newblock \emph{A Template for Scatter Search and Path Relinking}.
\newblock {\sl Lecture Notes in Computer Science} 1363, pp. 13--54

\bibitem {GLO2003}
Goldschmidt, Olivier, Laugier, Alexandre, Olinick, Eli V. (2003):
\newblock \emph{SONET/SDH Ring Assignment with Capacity Constraints}.
\newblock {\sl Discrete Appl. Math.} 129, pp. 99--128

\bibitem {GHLO2003}
Goldschmidt, Olivier, Hochbaum, Dorit S., Levin, Asaf, Olinick, Eli V. (2003):
\newblock \emph{The Sonet Edge-Partition Problem}.
\newblock {\sl Networks} 41, pp. 3--23

\bibitem {LSHK2000}
Lee, Youngho, Sherali, Hanif D., Han, Junghee, Kim, Seong-in (2000):
\newblock \emph{A Branch-and-Cut Algorithm for Solving an Intraring Synchronous Optical Network Design Problem}.
\newblock {\sl Networks} 35, pp. 223--232

\bibitem {CRT2001}
Cavique, Luís, Rego, César, Themido, Isabel (2001):
\newblock \emph{A Scatter Search Algorithm for the Maximum Clique Problem}.
\newblock {\sl Essays and Surveys in Metaheuristics Kluwer} pp. 227-244

\bibitem {HH2002}
Hamiez, Jean-Philippe, Hao, Jin-Kao (2002):
\newblock \emph{Scatter Search for Graph Coloring}.
\newblock {\sl Lecture Notes In Computer Science} 2310, pp. 195--213

\bibitem {P2009}
Pelleau, Marie (2009):
\newblock \emph{Sonet Network Design Problem}.

\end{thebibliography}

\end{document}